\documentclass{article}

\usepackage{arxiv}

\usepackage[utf8]{inputenc} 
\usepackage[T1]{fontenc}    
\usepackage{hyperref}       
\usepackage{url}            
\usepackage{booktabs}       
\usepackage{amsfonts}       
\usepackage{nicefrac}       
\usepackage{microtype}      
\usepackage{lipsum}		
\usepackage{graphicx}
\usepackage{natbib}
\usepackage{doi}

\usepackage{latexsym,amsmath,amssymb}

\usepackage{lipsum}
\usepackage{multicol}
\usepackage{graphicx}

\newtheorem{proposition}{Proposition}
\newtheorem{definition}{Definition}
\newtheorem{lemma}{Lemma}

\def\bfx{{\bf x}} \def\bfy{{\bf y}}
\def\bfr{{\bf r}}

\title{Bayesian Counterfactual Mean Embeddings and Off-Policy Evaluation}

\date{} 					

\author{Diego ~Martinez-Taboada\thanks{Work primarily done at the University of Oxford and finished at Carnegie Mellon University.} \\
	Department of Statistics \& Data Science\\
	Carnegie Mellon University\\
	\texttt{diegomar@andrew.cmu.edu} \\
	\And
	Dino ~Sejdinovic \\
	Department of Statistics\\
	University of Oxford \\
	\texttt{dino.sejdinovic@stats.ox.ac.uk} \\
}

\renewcommand{\undertitle}{Preprint. Under review.}
\renewcommand{\shorttitle}{\textit{arXiv} Template}

\hypersetup{
pdftitle={Bayesian Counterfactual Mean Embeddings and Off-Policy Evaluation},
pdfauthor={Diego ~Martinez-Taboada, Dino ~Sejdinovic},
}

\begin{document}
\maketitle

\begin{abstract}
	  The counterfactual distribution models the effect of the treatment in the untreated group. While most of the work focuses on the expected values of the treatment effect, one may be interested in the whole counterfactual distribution or other quantities associated to it. Building on the framework of Bayesian conditional mean embeddings, we propose a Bayesian approach for modeling the counterfactual distribution, which leads to quantifying the epistemic uncertainty about the distribution. The framework naturally extends to the setting where one observes multiple treatment effects (e.g. an intermediate effect after an interim period, and an ultimate treatment effect which is of main interest) and allows for additionally modelling uncertainty about the relationship of these effects. For such goal, we present three novel Bayesian methods to estimate the expectation of the ultimate treatment effect, when only noisy samples of the dependence between intermediate and ultimate effects are provided. These methods differ on the source of uncertainty considered and allow for combining two sources of data. Moreover, we generalize these ideas to the off-policy evaluation framework, which can be seen as an extension of the counterfactual estimation problem. We empirically explore the calibration of the algorithms in two different experimental settings which require data fusion, and illustrate the value of considering the uncertainty stemming from the two sources of data. 
\end{abstract}


\section{Introduction}

Uncertainty quantification (\cite{soize2017uncertainty}, \cite{psaros2022uncertainty}) is a cornerstone in modern statistics and machine learning, allowing for risk assessment and variability evaluation. Uncertainty is typically studied in two categories:  aleatoric uncertainty and epistemic uncertainty (\cite{sullivan2015introduction}). While the former refers to the intrinsic randomness of the studied object, the latter emerges due to the lack of knowledge about the object itself. Most of the literature focuses on modeling aleatoric uncertainty, namely by estimating the distribution or other statistic of a random object (\cite{monteiro2020stochastic}, \cite{wang2019aleatoric}, \cite{larranaga2001estimation}). However, quantification of both aleatoric and epistemic uncertainty is exploited in a wide range of settings, from Bayesian optimization to high-stakes applications such as assisted medical decision making (\cite{begoli2019need}) and chemistry (\cite{vishwakarma2021metrics}).

Epistemic uncertainty quantification is fundamental in scenarios suffering from covariate shift. The covariate shift (\cite{sugiyama2008direct}) emerges when the distribution in the test sample
\begin{align}
    \mathbb{P}^*_{X, Y}(x, y) = \mathbb{P}^*_{X}(x)\mathbb{P}_{Y | X}(y | x)
\end{align} 
differs in the marginal distribution of the covariates $X$ when compared to the train distribution
\begin{align}
    \mathbb{P}_{X, Y}(x, y) = \mathbb{P}_{X}(x)\mathbb{P}_{Y | X}(y | x).
\end{align}
Given the change in the distribution of the covariates from $\mathbb{P}_{X}$ to $\mathbb{P}^*_{X}$ may imply the evaluation of $\mathbb{P}_{Y | X}(y | x)$ in values of the covariate $x$ for which little information is available in the training data, epistemic uncertainty plays an important role in the assessment of the estimation (\cite{zhou2021bayesian}).

Covariate shift is inherent to counterfactual inference, which studies what
would have happened had some intervention been performed (\cite{pearl2003causality}, \cite{chernozhukov2013inference}): the counterfactual distribution evaluates the conditional treatment effect on the untreated group, thus clearly suffering from covariate shift. The positivity condition (\cite{westreich2010invited}) usually assumed in causal inference studies implies that such covariate shift is well-behaved, with the two covariate distributions having the same support. The violation of this assumption in real-life applications usually leads to erroneous conclusions (\cite{petersen2012diagnosing}, \cite{leger2022causal}). Furthermore, please note that the covariate shift problem assumes that $\mathbb{P}_{Y|X}$ remains invariant in the training and testing scenarios, which is equivalent to the exchangeability assumption (\cite{greenland1986identifiability}) frequently accepted in causal inference theory.

Closely related to counterfactual inference, off-policy evaluation (\cite{maei2010toward}) studies the effect of a target policy given data from a logging policy. Similarly to counterfactual distribution estimation, the off-policy evaluation problem also suffers from covariate shift. In this work, we propose a novel Bayesian approach for quantifying uncertainty in the two aforementioned problems. Our contributions are three folded:
\begin{itemize}
    \item We design a new Bayesian estimation of the counterfactual distribution.
    \item We design three new Bayesian approaches for modeling the expectation of a function of the counterfactual distribution, as well as a function of the off-policy evaluation distribution.
    \item We illustrate the approaches by running two experiments on synthetic data.
\end{itemize}

\section{Related work}

The Bayesian approaches presented in this work build on distribution representations through kernel embeddings (\cite{smola2007hilbert}). The concept of kernel mean embedding generalizes to conditional distributions by the so-called conditional mean embedding (\cite{song2009hilbert}). In terms of Bayesian procedures, \cite{flaxman_bayesian_kernel_embedding} proposed a Bayesian approach for learning mean embeddings, and \cite{bayesimp} generalized the Bayesian approach to conditional distributions. Such Bayesian learning of mean embeddings requires the concept of nuclear dominance (\cite{sample_path_rkhs}), which allows for defining a Gaussian process with trajectories in a reproducing kernel Hilbert space with probability 1.

In terms of counterfactual distribution estimation, it was first proposed to address the problem through quantile regression (\cite{melly2006estimation}). \cite{chernozhukov2013inference} extended the work by also considering duration and distribution regressions. Other approaches include density estimation (\cite{kennedy2021semiparametric}) or deep learning procedures (\cite{johansson2016learning}). 

Universal or distributional off-policy evaluation, whose aim is to estimate the whole distribution of the off-policy evaluation problem, has not received as much attention in the literature as the traditional off-policy evaluation, where only the expected value of such distribution is of interest (\cite{thomas2016data}, \cite{swaminathan2017off}). Recent research (\cite{OPE_risk_assessment}, \cite{universal_OPE}) proposed to estimate the cumulative density function of the distribution in order to later use it for plugin estimates. In terms of uncertainty quantification, \cite{conformal_off-policy_prediction} proposed a conformal approach to provide reliable predictive intervals.

A frequentist kernel mean embedding estimator for the counterfactual distribution estimation was proposed in \cite{muandet}, which also addressed off-policy evaluation through kernel embeddings. Kernel mean embedding-based methods are known for their performance in estimating the expectation of functions of the considered distribution (\cite{muandet2017kernel}), among other extended uses such as two sample tests (\cite{gretton2012kernel}). In contrast, alternative approaches that estimate the conditional density estimation for modeling the expectation of a function scale poorly with the dimension of the underlying space (\cite{grunewalder2012modelling}). 

For such reason, kernel mean embeddings have seen extensive use in two-staged estimators (\cite{singh2019kernel}, \cite{singh2021kernel}). Furthermore, conditional mean processes, introduced by \cite{chau2021deconditional}, study the integral of a Gaussian process with respect to a conditional distribution. Building on the Bayesian conditional mean embedding and the conditional mean process, \cite{bayesimp} proposed three algorithms for tackling a data fusion problem in a causal inference setting, which consider uncertainty derived from multiple datasets. Similar ideas were also explored in applications other than causal inference, with \cite{bayesian_sequential} exploiting Bayesian kernel embeddings for a sequential decision-making problem.

\section{Background}

\subsection{Counterfactual distribution and off-policy evaluation} \label{subsec:counter_ope}

The algorithms that will be presented in this work aim to address the counterfactual distribution estimation and off-policy evaluation problems. 

Counterfactual inference (\cite{johansson2016learning}, \cite{chernozhukov2013inference}) explores what would have happened had some intervention been performed, given that something else in fact occurred. Formally, let $T \in \{ 0,1 \}$ be a treatment and $\{ (X_0, Y_0) \}, \{ (X_1, Y_1) \}$ the observational data corresponding to the covariates and outcomes of two populations, respectively untreated and treated. The counterfactual distribution is defined as
\begin{equation}
\label{counterfactual_distribution}
\mathbb{P}_{Y \langle 0|1 \rangle} = \int \mathbb{P}_{Y_0|X_0}(y|x) d\mathbb{P}_{X_1}(x).
\end{equation}

Off-policy evaluation (OPE) arises in reinforcement learning settings (\cite{wang2017optimal}, \cite{OPE_risk_assessment}). It aims at understanding the effect of a target policy that has not been executed yet: it might be too expensive, unethical or impractical to implement it in order to draw conclusions (\cite{maei2010toward}, \cite{geist2014off}). Formally, the OPE problem is divided in three blocks: the space of context features $\mathcal{U}$, the space of actions $\mathcal{A}$, and the space of the rewards $\mathcal{R}$. While the characteristics of the subjects are encoded in $\mathbb{P}_U(u)$, distributions $\mathbb{P}_{A|U}(a|u)$ and $\mathbb{P}^*_{A|U}(a|u)$ model the logging policy (policy applied when the data is drawn) and the target policy respectively. We denote $\mathbb{P}_{U, A}(u, a) = \mathbb{P}_U(u) \mathbb{P}_{A|U}(a|u)$ and $\mathbb{P}^*_{U, A}(u, a) = \mathbb{P}_U(u) \mathbb{P}^*_{A|U}(a|u)$. Distribution $\mathbb{P}_{R|U, A}(r|u,a)$ models the response of the reward, which depends on both the policy and context features. The assessment of the policies is determined by the behaviour of the reward associated to such interventions. The goal of universal (or distributional) off-policy evaluation is to estimate
\begin{equation} \label{eq:ope}
    \mathbb{Q}^*(r) = \int_{\mathcal{U} \times \mathcal{A}} \mathbb{P}_{R|U, A}(r|u,a) d \mathbb{P}^*_{U, A}(u, a).
\end{equation}

There is a clear link between OPE and the counterfactual distribution estimation problem: $\mathbb{P}^*_{U, A}(u, a)$ plays the role of $\mathbb{P}_{X_1}(x)$. However, $\mathbb{P}_{X_1}(x)$ is fully estimated from data, while there is a given component $\mathbb{P}^*_{U|A}(u|a)$ in $\mathbb{P}^*_{U, A}(u, a)$ set by the policy makers. In fact, the counterfactual distribution problem could be embedded in the OPE framework, by considering a policy with no actions (i.e. $\mathcal{A} = \emptyset$). Furthermore, please note that both the counterfactual distribution and off-policy evaluation are clear examples where there exists a covariate shift (\cite{sugiyama2006mixture}, \cite{uehara}): distribution $\mathbb{P}_{Y_0|X_0}(y|x)$ is averaged over a new distribution $\mathbb{P}_{X_1}(x)$, and distribution $\mathbb{P}_{R|U, A}(r|u,a)$ is averaged over $\mathbb{P}^*_{U, A}(u, a)$ (distribution set by the policy maker).

\subsection{Off-policy and counterfactual evaluation with unmatched data} \label{subsec:unmatched}

In the usual OPE and counterfactual estimation settings only the expectation $\mathbb{E}[R]$, where $R \sim \mathbb{P}_{Y \langle 0|1 \rangle}$ or $R \sim \mathbb{Q}^*$, is of interest. However, one may be interested in the expectation of a function $f(R)$, instead of $R$ itself. In various fields of application, it is usual to study a function of the outcome, rather than the outcome itself. For instance, the conditional value at risk (CVaR), which has been extensively used for quantifying financial risks (\cite{rockafellar2000optimization}, \cite{zhu}), takes $f(R) = R 1_{R \leq r_\alpha}$. 

In more general settings, the function $f$ may be itself unknown. Noisy data of the form
\begin{equation}
    (r_i, f(r_i) + \xi_i)_{i = 1}^M
\end{equation}
may be the only available information on such dependency. For example, consider a clinical trial where the effect of an intervention is measured in terms of chest X-rays, while a data set might have been already collected classifying chest X-rays from healthy patients or patients with lung cancer. In this scenario, $r_i$ would represent the X-rays and $f(r_i)$ the classification (healthy or unhealthy patient) of such X-ray.

Please note that even if we assume $\xi_i$ to be zero (i.e. observations with no noise), the expectation $\eta = \mathbb{E}[f(R)]$ cannot be expressed in terms of the expectation $\mathbb{E}[R]$. Furthermore, this scenario allows for considering multiple sources of data. Algorithms that handle unmatched data sets may be of interest in heterogeneous fields of application such as the aforementioned clinical example, with the so called data fusion problem (\cite{meng2020survey}) receiving growing attention recently.

\subsection{RKHS and Bayesian mean embeddings}

The framework of Bayesian mean embeddings (\cite{flaxman_bayesian_kernel_embedding}) is the cornerstone of the algorithms presented in this work, used for modeling the whole counterfactual distribution. In this section, we present the theoretical background related to Reproducing Kernel Hilbert Spaces (RKHS), conditional mean embeddings, and Bayesian approaches relative to the Bayesian conditional mean embedding. 

\begin{definition}[Reproducing Kernel Hilbert Space]
Let $\mathcal{X}$ be a non-empty set and let $\mathcal{H}$ be a Hilbert space of functions $f: \mathcal{X} \to \mathbb{R}$ with inner product $\langle \cdot, \cdot \rangle_\mathcal{H}$. A function $k: \mathcal{X} \times \mathcal{X} \to \mathbb{R}$ is called a reproducing kernel of $\mathcal{H}$ if it satisfies 
\begin{itemize}
    \item $k(\cdot, x) \in \mathcal{H}\quad \forall x \in \mathcal{X}$,
    \item (the reproducing property) $\langle f, k(\cdot, x) \rangle_\mathcal{H} = f(x) \quad \forall x \in \mathcal{X}, \forall f \in \mathcal{H}.$
\end{itemize}
If $\mathcal{H}$ has a reproducing kernel, then it is called reproducing kernel Hilbert space (RKHS).
\end{definition}

We refer to the considered kernel associated to set $\mathcal{X}$ as $k_x$.  An element $k(\cdot, x) \in \mathcal{H}$ is also referred as $\phi_x(x)$. Furthermore, given a data vector $\textbf{x} = [x_1, ..., x_n]^T$, we define the feature matrices $\Phi_{\bfx} := [\phi_x(x_1), ..., \phi_x(x_n)]$. We denote the Gram matrix as $K_{\bfx \bfx} = \Phi_{\bfx}^T \Phi_{\bfx}$, while $k_{x \bfx} = [k_x(x, x_1), ..., k_x(x, x_n)]$ is the vector of evaluations. Moreover, we denote $\Phi_{\bfx}(x) = k_{x \bfx}^T$. The notation is analogously used for other sets and variables considered.

The Kernel Mean Embedding (KME), which builds on the concept of RKHS, maps distributions to elements in the respective Hilbert space. 
 
\begin{definition}[Kernel Mean Embedding (KME)]
Let $\mathcal{P}$ be the set of all probability measures on a measurable space $(\mathcal{X}, \mathcal{B}_\mathcal{X})$, and $k: \mathcal{X} \times \mathcal{X}$ a reproducing kernel with associated RKHS $\mathcal{H}$ such that $sup_{x \in \mathcal{X}} k(x,x) < \infty$. The kernel mean embedding (KME) of $\mathbb{P} \in \mathcal{P}$ with respect to $k$ is defined as the following Bochner integral
$$\mu: \mathcal{P} \to \mathcal{H}, \quad \mathbb{P} \to \mu_\mathbb{P}:= \int k(\cdot, x) d\mathbb{P}(x).$$
\end{definition}

If the kernel considered is \textit{characteristic} (which is the case for frequently used kernels such as the RBF or Matern kernels), then $\mu$ is injective and hence the KME allows for distribution representation (\cite{fukumizu2007kernel}). Furthermore, kernel mean embeddings may also be used to find the expectation of a function in the RKHS as stated in the following lemma.

\begin{lemma} \label{lemma:kme_mean}
The kernel mean embedding of $\mathbb{P}$ maps functions $f \in \mathcal{H}$ to their mean with respect to $\mathbb{P}$ through the inner product:
\begin{equation} 
    \langle \mu_{\mathbb{P}}, f \rangle_\mathcal{H} = 
    \mathbb{E}_{X \sim \mathbb{P} }[f(X)].
\end{equation}
\end{lemma}

Conditional mean embeddings (\cite{song2009hilbert}) extend the concept of kernel mean embeddings to conditional distributions. The conditional mean embedding operator is a Hilbert-Schimdt operator $\mathcal{C}_{Y|X}: \mathcal{H}_{k_x} \to \mathcal{H}_{k_y}$ satisfying  $\mu_{Y | X = x} = \mathcal{C}_{Y|X} \phi_x(x)$, where $ \mathcal{C}_{Y|X} := \mathcal{C}_{YX} \mathcal{C}_{XX}^{-1}$, $\mathcal{C}_{YX} := \mathbb{E}_{Y, X}[\phi_y(y) \otimes \phi_x(x)]$ and $\mathcal{C}_{XX} := \mathbb{E}_{X, X}[\phi_x(x) \otimes \phi_x(x)]$. Given a dataset $\{ \bfx, \bfy \}$, a sample estimator may be defined as 
\begin{equation} \label{eq:conditional_mean_embedding}
    \hat{\mathcal{C}}_{Y|X} = \Phi_{\bfy}^T(K_{\bfx \bfx} + \lambda I)^{-1} \Phi_{\bfx},
\end{equation}
where $\lambda$ is the regularization term. As exhibited in \cite{cme_regressor}, such sample estimator may be interpreted as a kernel ridge regression. Please note that the regression task avoids the computation of density estimation.

A Bayesian learning framework on kernel mean embeddings was proposed in \cite{flaxman_bayesian_kernel_embedding}, followed by a generalization to conditional mean embeddings exhibited in \cite{bayesimp}. \cite{bayesimp} proposed to model the conditional mean embedding $\mu_{Y | X = x}(y)$ as a Gaussian process $\mu_{gp}(x, y)$, with prior $\mu_{gp} \sim GP(0, k_x \otimes r_y)$. For $\mu_{gp}(x, \cdot)$ to model $\mu_{Y|X = x}(\cdot)$, the paths of $\mu_{gp}(x, \cdot)$ should live in the RKHS associated to kernel $k$ almost surely. If $r_y = k_y$, then the paths live outside the RKHS with probability 1 (\cite{sample_path_rkhs}). However, if the prior covariate kernel is defined as a nuclear dominant kernel over $k_y$, the paths of $\mu_{gp}(x, y)$ live almost surely within the RKHS (\cite{sample_path_rkhs}). Similarly to \cite{flaxman_bayesian_kernel_embedding}, we choose $r_y$ to be the convolution of the original kernel with itself. Hence the prior over $\mu_{gp}$ is
\begin{equation}
\mu_{gp} \sim GP(0, k_x \otimes r_y),
\end{equation}
\begin{equation}
r_y(y_1, y_2) := \int k_y(y_1, u) k_y(u, y_2) \nu (du),
\end{equation}
where $\nu$ is a finite measure on $\mathcal{Y}$. We then define the function-valued regression 
\begin{equation}
    \phi_y(y_i) = \mu_{gp}(x_i, \cdot) + \lambda^{1/2} \epsilon_i,
\end{equation}
where $\epsilon_i \stackrel{iid}{\sim} GP(0, r)$ are noise functions. Closed forms are available for the posterior mean and covariance of $\mu_{gp}$, the marginal likelihood, and nuclear dominant kernels for specific choices of kernels; we refer to \cite{bayesimp} for their derivations. We exhibit the posterior mean and covariance of $\mu_{gp}$ for completeness in the following proposition.

\begin{proposition}[Bayesian Conditional Mean Embedding (BayesCME)] \label{proposition: BL-CME}
Posterior distribution of CME $F(x, y)=\mu_{Y|X=x}(y)$  given observations $\{ \bfx, \bfy \}$ is a GP with the following mean and covariance:
\begin{align}
    m_\mu((x, y)) &= k_{x\bfx}(K_{\bfx\bfx} + \lambda I)^{-1}K_{\bfy\bfy}R_{\bfy\bfy}^{-1}r_{\bfy y}, \\
    \kappa_\mu((x, y), (x', y')) &= k_{xx'}r_{y, y'} - k_{x\bfx}(K_{\bfx\bfx} + \lambda I)^{-1}k_{\bfx x'}r_{y\bfy}R_{\bfy\bfy}^{-1}r_{\bfy y'}.
\end{align}
\end{proposition}

\subsection{Counterfactual Mean Embedding}

We present the Counterfactual Mean Embedding (CFME), introduced in \cite{muandet}, which represents the counterfactual distribution in terms of the kernel mean embedding. 

\begin{definition}[Counterfactual mean embedding (CFME)]
The kernel mean embedding of the counterfactual distribution is called counterfactual mean embedding and it is defined as follows:
\begin{equation}
    \mu_{Y \langle 0|1 \rangle} := \mu_{\mathbb{P}_{Y \langle 0|1 \rangle}} = \int k(\cdot, y) d\mathbb{P}_{Y \langle 0|1 \rangle}(y).
\end{equation} 
\end{definition}

Kernel $k$ is usually chosen to be characteristic, so $\mu_{Y \langle 0|1 \rangle}$ serves as a representation of the counterfactual distribution. Please note that 
\begin{equation}
    \mu_{Y \langle 0|1 \rangle} = \int k(\cdot, y) d\mathbb{P}_{Y \langle 0|1 \rangle}(y) = \int \mu_{Y_0 | X_0 = x} d \mathbb{P}_{X_1}(x),
\end{equation}
where $\mu_{Y_0 | X_0 = x} = \int k(\cdot, y) d\mathbb{P}_{Y_0|X_0 = x}(y)$ is the conditional mean embedding.

Consequently, an empirical estimator of the counterfactual mean embedding may be derived as follows. Suppose two independent samples 
\begin{equation}
    \{ (x^0_i, y_i)_{i = 1}^n \} \stackrel{iid}{\sim} \mathbb{P}_{X_0 Y_0}(x, y), \quad x_1^1, ..., x_m^1 \stackrel{iid}{\sim} \mathbb{P}_{X_1}(x)
\end{equation}
are given. Provided $\mu_{Y \langle 0|1 \rangle} = \int \mu_{Y_0 | X_0 = x} d \mathbb{P}_{X_1}(x)$, the empirical estimator may be defined as 
\begin{equation} \label{eq:empirical_counterfactual}
\hat{\mu}_{Y \langle 0|1 \rangle} := \frac{1}{m} \sum_{j = 1}^m \hat{\mu}_{Y_0 | X_0 = x_j^1},
\end{equation}
where $\hat{\mu}_{Y_0 | X_0 = x_j^1} = \hat{\mathcal{C}}_{Y|X} \phi_x(x_j^1) = \Phi_{\bfy}(K_{\bfx \bfx} + \lambda I)^{-1} \Phi_x(x_j)$, $\bfx = [x_1^0, ..., x_n^0]^T$, and $\bfy = [y_1, ..., y_n]^T$. We refer to \cite{muandet} for a convergence analysis of such estimator.

\section{A Bayesian approach for the counterfactual distribution}

We are now ready to introduce the Bayesian counterfactual mean embedding (BayesCFME).  It exploits the same idea as the frequentist counterfactual mean embedding (CFME): the addition of the conditional mean embeddings given another sample of points to be conditioned on. However, the BayesCFME builds on the Bayesian conditional mean embedding, thus providing with uncertainty estimates (which, as we will see, will be crucial in the unmatched data setting). The closed-form solution exhibited in the succeeding proposition follows from combining Proposition \ref{proposition: BL-CME} and the counterfactual mean embedding empirical estimator defined in Equation \ref{eq:empirical_counterfactual}.

\begin{proposition}[Bayesian Counterfactual Mean Embedding (BayesCFME)] \label{prop_BayesCFME}
The posterior distribution of the counterfactual mean embedding $\mu_{Y \langle 0|1 \rangle}$ given independent samples $\{ (x_i, y_i)_{i = 1}^n \} \stackrel{iid}{\sim} \mathbb{P}_{X Y}(x, y)$ and $x_1', ..., x_m' \stackrel{iid}{\sim} \mathbb{P}_{X'}(x)$ is a GP with the following mean and covariance:
\begin{align}
    m(y) &= \frac{1}{m}1_m^{\top} K_{\bf{x'}\bfx}(K_{\bfx\bfx} + \lambda I)^{-1}K_{\bfy\bfy}R_{\bfy\bfy}^{-1}r_{\bfy y}, \\
    \kappa( y, y') &= \frac{1}{m^2} 1_m^{\top} \left( K_{\bf{x'}\bf{x'}}r_{y, y'} - K_{\bf{x'}\bfx}(K_{\bfx\bfx} + \lambda I)^{-1}K_{\bfx \bf{x'}}r_{y\bfy}R_{\bfy\bfy}^{-1}r_{\bfy y'}) \right)1_m.
\end{align}
\end{proposition}

\begin{figure}[t]
    \centering 
    \includegraphics[width=\linewidth]{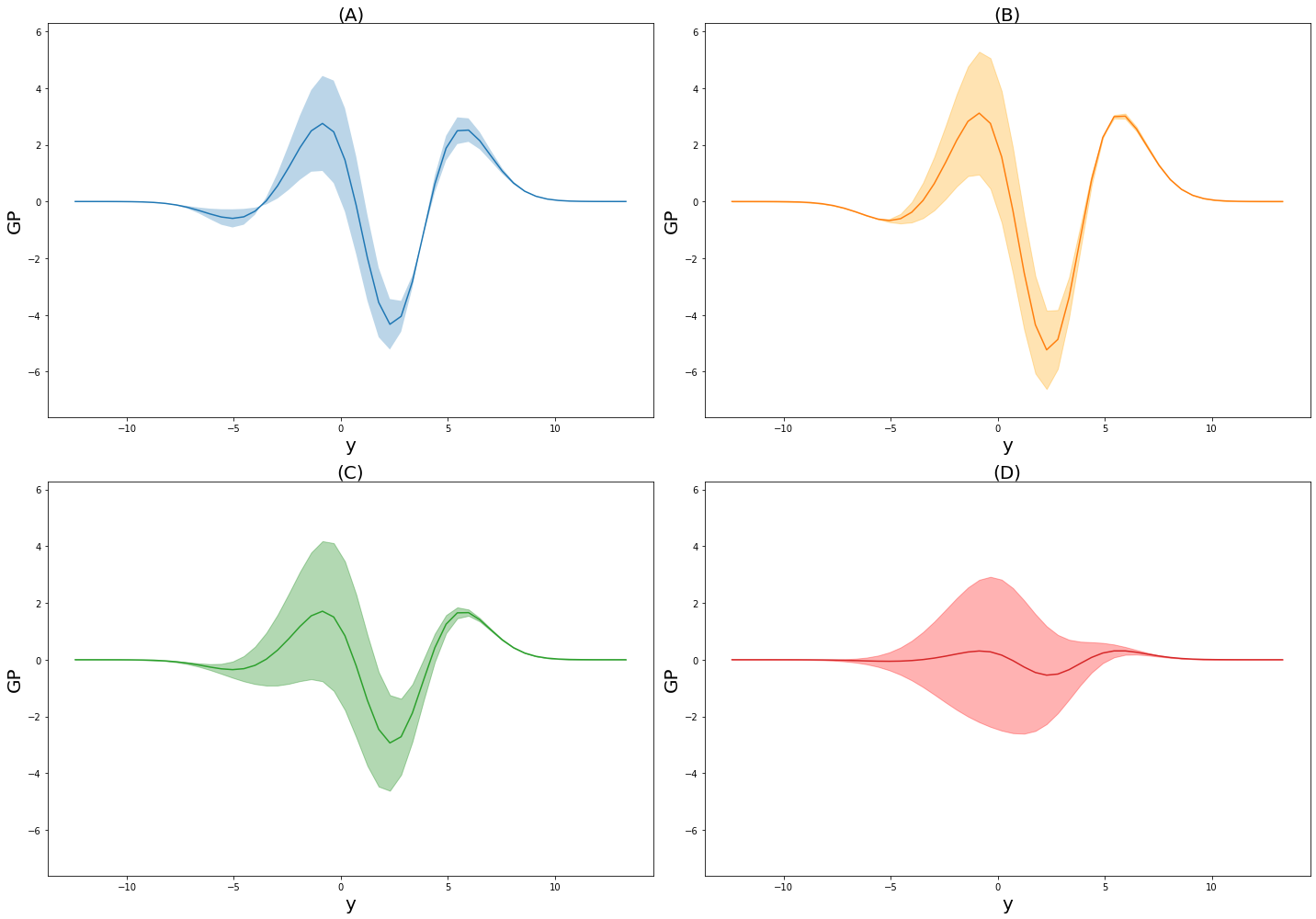} 
    \caption{Illustration of the GP defining the posterior distribution of an example of a counterfactual mean embedding considered. The sample $\bf{x}'$ is defined as $\bf{x} + $ shift, where (A) shift = 5, (B) shift = 7, (C) shift = 9, (D) shift = 11.}
    \label{fig:covariate_shift_2}
\end{figure}

Please note the impact of the distributional shift $\mathbb{P}_{X_1}$ through $K_{\bfx' \bfx'}$ and $K_{\bfx' \bfx}$. For example, if the distributional shift is given by $f_{X_1}(x) = f_{X_0}(x + a)$, then we expect $K_{\bfx' \bfx} \to 0$ as $a \to \infty$. However, the behaviour of the entries of $K_{\bfx' \bfx'}$ is not affected by such distributional shift and so the covariance goes to the first term of $ \kappa( y, y')$. Similarly, $K_{\bfx' \bfx} \to 0$ implies $m(y) \to 0$ for a fixed $y \in \mathcal{Y}$. The distributional shift causes the mean of the posterior distribution of the GP to go to zero. Figure \ref{fig:covariate_shift_2} illustrates the Bayesian counterfactual mean embedding for different distributional shifts. Please note that, the greater the shift, the more uncertainty raises. Furthermore, the GP posterior mean goes to zero as we consider points further away from the original sample: the information of the conditional distribution is reduced as we distance from the original points conditioned on.

\begin{figure}[t]
    \centering 
    \includegraphics[width=\linewidth]{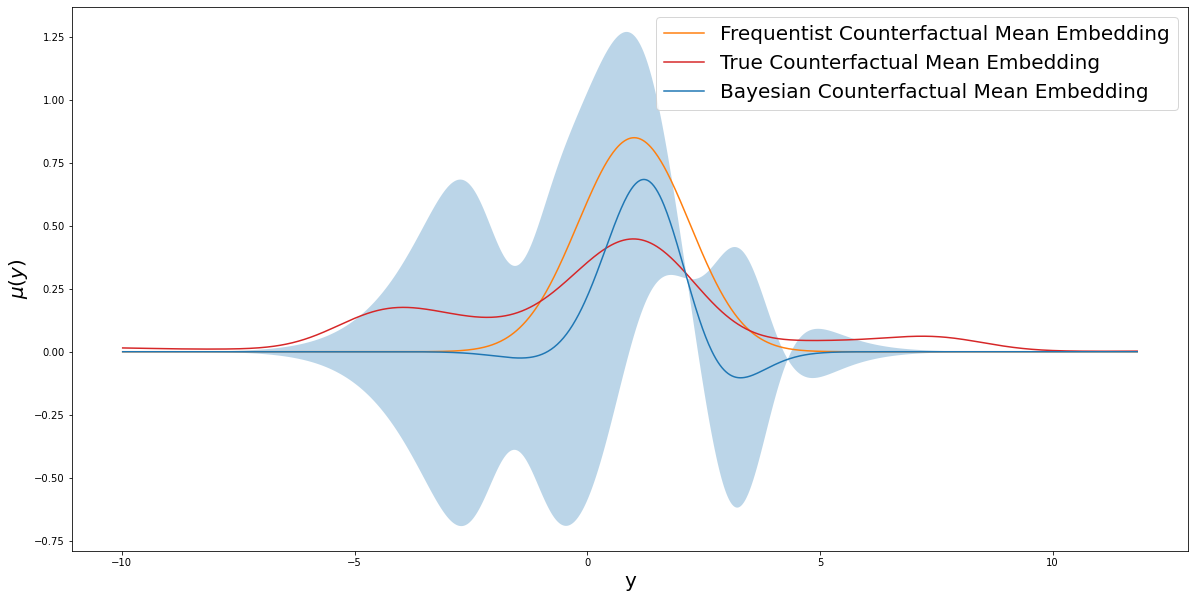} 
    \caption{Illustration of the true counterfactual mean embedding, the frequentist counterfactual mean embedding considered in \cite{muandet2017kernel}, and the Bayesian counterfactual mean embedding introduced in this work.}
    \label{fig:freq_bayesian}
\end{figure}

In contrast to CFME, BayesCFME provides with uncertainty estimates. The mean embedding itself encloses the aleatoric uncertainty of the counterfactual i.e. the counterfactual distribution itself. The uncertainty estimates of the BayesCFME model the epistemic uncertainty, as they are obtained due to lack of information. Figure \ref{fig:freq_bayesian} illustrates the comparison between the frequentist counterfactual mean embedding with kernel $K_y$ and the Bayesian counterfactual mean embedding with prior nuclear dominant kernel $R_y$. The introduction of the nuclear dominant kernel in the Bayesian approach results in the frequentist counterfactual mean embedding being different to the mean of the Bayesian counterfactual mean embedding.

One may be interested in the BayesCFME for a variety reasons. Firstly, considering a Bayesian approach of the CFME for more than one group allows for a Bayesian kernel two-sample testing (\cite{zhang2022bayesian}). Furthermore, the BayesCFME may be used for calculating the expectation of functions with respect to $\mathbb{P}_{Y \langle 0|1 \rangle}$, given that they belong to the respective RKHS. The uncertainty estimates provided by BayesCFME allow for uncertainty on the predicted expectation. The idea is analogous to the one used in \textit{Bayesian Counterfactual Mean Embedding}, which will be discussed in the next section (but with no need to estimate the function of interest). 

Please note that BayesCFME may also be applied to the off-policy evaluation problem, with the only difference that the sample $\{x_i\}_{i = 1}^n = \{u_i, a_i\}_{i = 1}^n$ from Proposition \ref{prop_BayesCFME} is partially generated, given that we ought to simulate $a_i | u_i$ from the proposed policy.

\section{A Bayesian approach for the off-policy evaluation problem} \label{sec:algo}

We now propose three estimators for the problem introduced in Subsection \ref{subsec:unmatched}, based on the ideas presented in \cite{bayesimp}. Similarly to \cite{bayesimp}, we propose two-staged estimators that take into account uncertainties stemming from different data sources. However, the main difference is that the algorithms presented in \cite{bayesimp} are designed to estimate the effect of an intervention, while we address a counterfactual problem.  

Let's suppose that we are given samples $D_1 = \{ (x_i, r_i) \}_{i = 1}^N \sim \mathbb{P}_{X_0, Y_0}$, $D_2 = \{ (\tilde{r_j}, f(\tilde{r_j}) + \xi_j) \}_{j = 1}^M$, $D_3 = \{ x'_l \}_{l = 1}^L \sim \mathbb{P}_{X_1}$, and the goal is to estimate $\mathbb{E}[f(R)]$, where $R \sim \mathbb{P}_{Y \langle 0|1 \rangle}$. For simplicity of notation, we denote $y_j = f(\tilde{r_j}) + \xi_j$. Alternatively, we can define the analogous off-policy problem where $D_1 = \{ (u_i, a_i, r_i) \}_{i = 1}^N \sim \mathbb{P}_{U, A, R}$, $D_2$ is defined in the same way, and $D_3 = \{ (u'_l, a'_l) \}_{l = 1}^L \sim \mathbb{P}^*$, which is the target policy. In this case, part of $D_3$ is to be generated by the policy maker. However, we can refer to $(u, a)$ as $x$, given that variables $u$ and $a$ play the same role from a mathematical point of view. 

Following a structure similar to \cite{bayesimp}, we propose  two-staged estimators. On the one hand, we estimate $\mathbb{P}_R$ (distributional OPE) based on $D_1$ and $D_3$. On the other hand, we estimate $f(r) := \mathbb{E}_p[Y | r]$ given $D_2$. We consider three alternatives in this framework, which differ on whether a GP or a frequentist approach is used for the estimation of $\mathbb{P}_R$ and $f$:
\begin{itemize}
    \item \textbf{Counterfactual Mean Process (CFMP)}: The expectation $f(r)$ is trained as a GP and $\mu_{R \langle 0, 1 \rangle}$ is modeled by CFME.
    
    \item \textbf{Bayesian Regressing Counterfactual Mean Embedding (BayesRCFME)}: The expectation $f(r)$ is modeled as a real-valued kernel ridge regression and and $\mu_{R \langle 0, 1 \rangle}$ is trained as a BayesCFME.
	
    \item \textbf{Bayesian Counterfactual Mean Process (BayesCFMP)}: The expectation $f(r)$ is trained as a GP and and $\mu_{R \langle 0, 1 \rangle}$ is trained as a BayesCFME.
    
\end{itemize}

While the first two options fail to capture the uncertainty inherited by $D_1$ and $D_2$ respectively, \textit{BayesCFMP} encompasses uncertainty derived from both data sets. Similarly to the algorithms presented \cite{bayesimp}, closed form solutions exist for the three approaches introduced thereof. While the algorithms exhibited in \cite{bayesimp} have to account for an adjustment set, the proposed methods have to average the mean embeddings over $D_3$, hence using the counterfactual mean embedding. However, the proofs of the propositions are analogous. We state the closed forms in here for completeness, and we refer \cite{bayesimp} for their derivations. We start by introducing the closed form solution of \textit{CFMP}.

\begin{proposition} [Counterfactual Mean Process (CFMP)]
Let $k_u, k_a, k_r$ be the kernels associated to variables $u, a, k$ respectively and $D_1 = \{ (u_i, a_i, r_i) \}_{i = 1}^{N} = \{ (x_i, r_i) \}_{i = 1}^{N}$, $D_2 = \{ (\tilde{r}_j, y_j) \}_{j = 1}^M$ two unmatched datasets . Furthermore, let $D_3 = \{ (u_i', a_i') \}_{i = 1}^{L} = \{ x' \}_{i = 1}^{L}$ be sampled from $\mathbb{P}^*$. We denote $\Phi(x) = \Phi(s, a) = \Phi_{\bf{s}}^T k_s(s, \cdot) \odot \Phi_{\bf{a}}^T k_a(a, \cdot)$. If $f$ is the posterior GP learnt form $D_2$, then $\eta = \int f(r) q^*(r) \sim \mathcal{N}(\mu_1, \sigma^2_1)$ where: 
\begin{align}
    \mu_1 &= \langle \hat{\mu}_{R \langle 0, 1 \rangle}, m_f \rangle = \frac{1}{L}1_L^{\top} K_{\bf{x'}\bfx} (K + \lambda I)^{-1} K_{\bfr \tilde{\bfr}}  (K_{\tilde{\bfr} \tilde{\bfr} } + \lambda_f I)^{-1} y, \\
    \sigma^2_1 &= \mu_{Y \langle 0|1 \rangle}^{\top}  \mu_{Y \langle 0|1 \rangle} - \mu_{Y \langle 0|1 \rangle}^{\top} \Phi_{\tilde{\bfr}} (K_{\tilde{\bfr} \tilde{\bfr} } + \lambda I)^{-1} \Phi_{\tilde{\bfr}}^{\top} \mu_{Y \langle 0|1 \rangle} \\ &= \frac{1}{L^2}1_L^{\top} K_{\bf{x'}\bfx} (K + \lambda I)^{-1} \tilde{K}_{\bfr \bfr} (K + \lambda I)^{-1} K_{\bfx \bf{x'}} 1_L, 
\end{align}
where $K_{\tilde{\bfr} \bfr} = \Phi_{\tilde{\bfr}}^\top \Phi_{\bfr}$, $m_f$ and $\tilde{K}_{\bfr \bfr}$ are the posterior mean function and covariance of $f$ evaluated at $r$. $\lambda > 0$ is the regularization parameter and $\lambda_f$ is the noise term for the GP. 
\end{proposition}

In CFMP, the GP $f$ is not required to be in any RKHS. In contrast, nuclear dominance is needed in the next two approaches. They exploit the fact that kernel mean embeddings may be used to compute the expectation of $f$ through the inner product, but $f$ ought to be in the respective RKHS (Lemma \ref{lemma:kme_mean}). The closed form of the BayesRCFME expressions follow. 

\begin{proposition} [Bayesian Regressing Counterfactual Mean Embedding (BayesRCFME)] \label{proposition:bayescfme}
Let $k_u$, $k_a$, $k_r$ be the kernels associated to variables $u, a, k$ respectively and $D_1 = \{ (u_i, a_i, r_i) \}_{i = 1}^{N} = \{ (x_i, r_i) \}_{i = 1}^{N}$, $D_2 = \{ (\tilde{r}_j, y_j) \}_{j = 1}^M$ two unmatched datasets. Let $D_3 = \{ (u_i', a_i') \}_{i = 1}^{L} = \{ x' \}_{i = 1}^{L}$ be sampled from $\mathbb{P}^*$. We denote $\Phi(x) = \Phi(s, a) = \Phi_{\bf{s}}^T k_s(s, \cdot) \odot \Phi_{\bf{a}}^T k_a(a, \cdot)$. If $f$ is a KRR learnt form $D_2$ and $\mu_{Y \langle 0|1 \rangle}$ modelled as a V-GP using $D_1$, then $\eta = \langle f, \mu_{Y \langle 0|1 \rangle} \rangle \sim \mathcal{N}(\mu_2, \sigma^2_2)$, where
\begin{align}
    \mu_2 &= \frac{1}{L}1_L^{\top} K_{\bf{x'}\bfx}(K_{\bfx\bfx} + \lambda I)^{-1}K_{\bfr\bfr}R_{\bfr\bfr}^{-1}R_{\bfr \bf\tilde{r}}A, \\
    \sigma^2_2 &= B \frac{1}{L^2} 1_L^{\top} K_{\bf{x'}\bf{x'}}1_L - C \frac{1}{L^2} 1_L^{\top} K_{\bf{x'}\bfx}(K_{\bfx\bfx} + \lambda I)^{-1} K_{\bfx \bf{x'}} 1_L,
\end{align}
where $A = (K_{\bf\tilde{r} \bf\tilde{r}} + \lambda_f I)^{-1} y$,  $B = A^T R_{\bf\tilde{r} \bf\tilde{r}} A$, and $C = A^T R_{\bf\tilde{r} \bfr} R_{\bfr \bfr}^{-1} R_{\bfr \bf\tilde{r}} A$.
\end{proposition}

Thirdly, we present the closed form solutions of BayesCFMP, which accounts for both sources of uncertainty.

\begin{proposition} [Bayesian Counterfactual Mean Process (BayesCFMP)] \label{proposition:bayescfmp}
Let $k_u$, $k_a$, $k_r$ be the kernels associated to variables $u, a, k$ respectively and $D_1 = \{ (u_i, a_i, r_i) \}_{i = 1}^{N} = \{ (x_i, r_i) \}_{i = 1}^{N}$, $D_2 = \{ (\tilde{r}_j, y_j) \}_{j = 1}^M$ two unmatched datasets. Let $D_3 = \{ (u_i', a_i') \}_{i = 1}^{L} = \{ x' \}_{i = 1}^{L}$ be sampled from $\mathbb{P}^*$. We denote $\Phi(x) = \Phi(s, a) = \Phi_{\bf{s}}^T k_s(s, \cdot) \odot \Phi_{\bf{a}}^T k_a(a, \cdot)$, and $\hat{\bfr} = (\bfr, \tilde{\bfr})$. If $f$ and $\mu_{Y \langle 0|1 \rangle}$ are modeled as GP, then $\eta = \langle f, \mu_{Y \langle 0|1 \rangle} \rangle$ has the following mean $\mu_3$ and variance $\sigma^2_3$:
\begin{align}
    \mu_3 &= E K_{\bfr \bf\hat{r}} K_{\bf\hat{r} \bf\hat{r}}^{-1} R_{\bf\hat{r} \bf\tilde{r}} (R_{\bf\tilde{r} \bf\tilde{r}} + \lambda_f I)^{-1} y, \\
    \sigma^2_3 &= E \Theta_1^{\top} \bar{R}_{\bf\hat{r} \bf\hat{r}} \Theta_1 E^{\top} + \Theta_2^{(a)} F - \Theta_2^{(b)} G  + \Theta_3^{(a)} F - \Theta_3^{(b)} G, 
    \end{align}
where 
$E = \frac{1}{L}1_L^{\top} K_{\bf{x'}\bfx}(K_{\bfx\bfx} + \lambda I)^{-1}$, 
$F = \frac{1}{L^2} 1_L^{\top} K_{\bf{x'}\bf{x'}}1_L$, 
$G = \frac{1}{L^2} 1_L^{\top} K_{\bf{x'}\bfx}(K_{\bfx\bfx} + \lambda I)^{-1} K_{\bfx \bf{x'}} 1_L$,
$\Theta_1 = K_{\bf\hat{r} \bf\hat{r}}^{-1} R_{\bf\hat{r} \bfr} R_{\bfr \bfr}^{-1} K_{\bfr \bfr}$, 
$ \Theta_2^{(a)} = \Theta_4^{\top} R_{\bf\hat{r} \bf\hat{r}} \Theta_4$,
$\Theta_2^{(b)} = \Theta_4^{\top} R_{\bf\hat{r} \bfr} R_{\bfr \bfr}^{-1} R_{\bfr \bf\hat{r}} \Theta_4$,
$\Theta_3^{(a)} = tr(K_{\bf\hat{r} \bf\hat{r}}^{-1} R_{\bf\hat{r} \bf\hat{r}} K_{\bf\hat{r} \bf\hat{r}}^{-1} \bar{R}_{\bf\hat{r} \bf\hat{r}})$, 
$\Theta_3^{(b)} = tr(R_{\bf\hat{r} \bfr} R_{\bfr \bfr}^{-1} R_{\bfr \bf\hat{r}} K_{\bf\hat{r} \bf\hat{r}}^{-1} \bar{R}_{\bf\hat{r} \bf\hat{r}} K_{\bf\hat{r} \bf\hat{r}}^{-1})$, 
$ \Theta_4 = K_{\bf\hat{r} \bf\hat{r}}^{-1} R_{\bf\hat{r} \bf\tilde{r}} (K_{\bf\tilde{r} \bf\tilde{r}} + \lambda_f)^{-1} y$,
.
$\bar{R}_{\bf\hat{r} \bf\hat{r}}$ is the posterior covariance of f evaluated at $\hat{\bfr}$.

\end{proposition}

\section{Experiments}

\begin{figure}[t]
    \centering 
    \includegraphics[width=\linewidth]{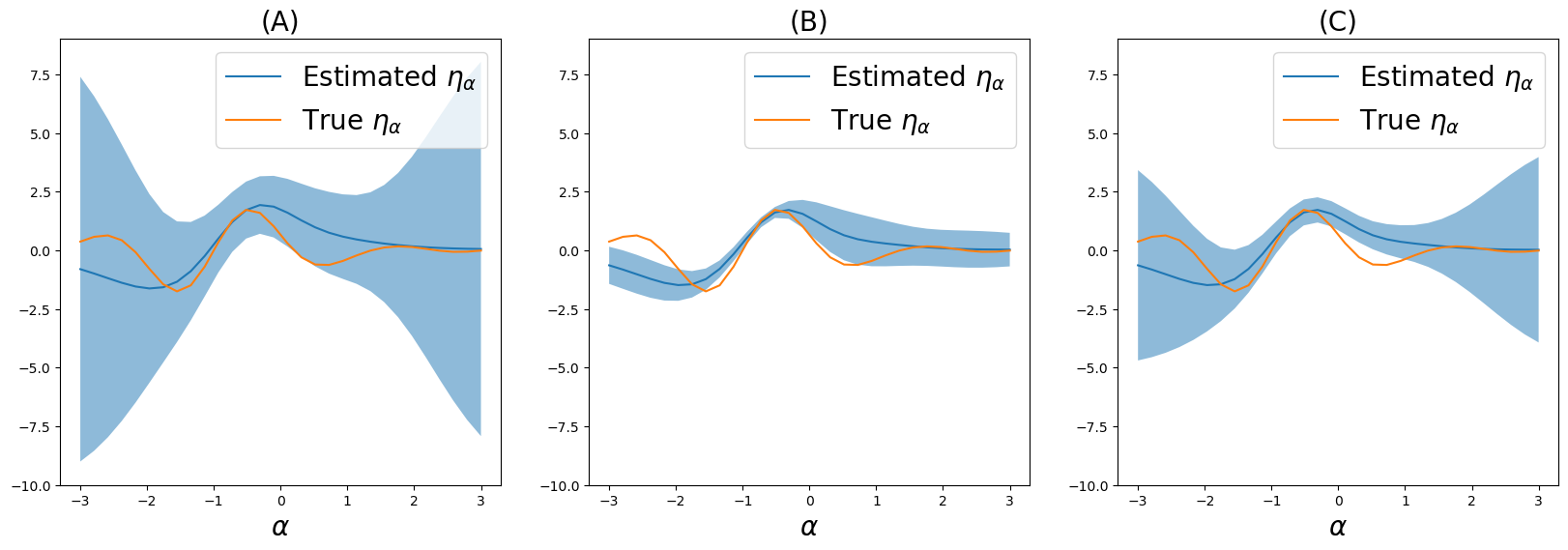} 
    \caption{Illustration of the experiment conducted on setting A. The mean and uncertainty estimates (95\% CI) of $\eta_\alpha$ are displayed for (A) BayesCFMP, (B) CFMP, (C) BayesCFME.}
    \label{fig:experiment}
\end{figure}

In order to illustrate the performance of the proposed algorithms, two simulated experimental settings are considered on different off-policy evaluation problems (please note that the counterfactual estimation problem presented in Section \ref{sec:algo} could be understood as a specific example of the OPE problem as stated in Subsection \ref{subsec:counter_ope}). For each of the examples, we work with two generated data sets $D_1 = \{ (u_i, a_i, r_i) \}_{i = 1}^N$ (collected from the logging policy) and $D_2 = \{ (\tilde{r}_j, y_j) \}_{j = 1}^M$. We then propose a set of new policies $\mathbb{P}_{A | U}^{\alpha}(a|u)$, where $\alpha \in \Lambda \subset \mathbb{R}$. The goal is to estimate $\eta_\alpha$ (ultimate effect of policy $\mathbb{P}^\alpha$) and to provide with uncertainty estimates accounting for the reliability of the measurement. 

\begin{figure}[b!]
    \centering 
    \includegraphics[width=\linewidth]{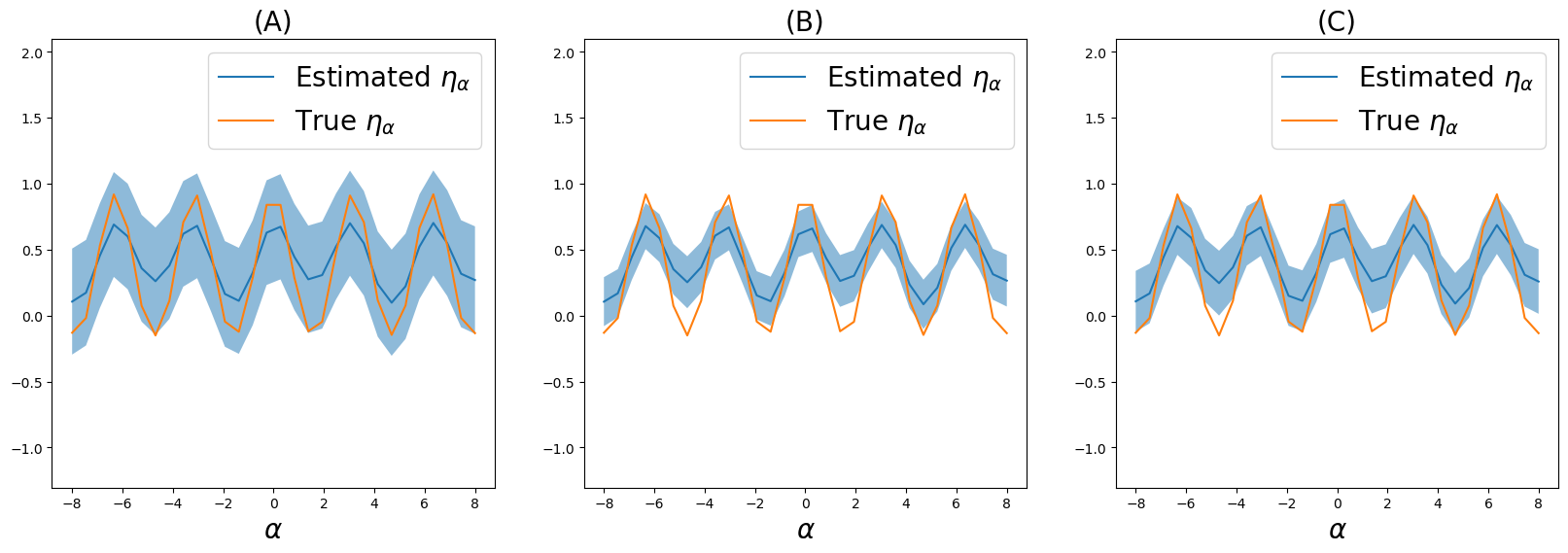} 
    \caption{Illustration of the experiment conducted on setting B. The mean and uncertainty estimates (95\% CI) of $\eta_\alpha$ are displayed for (A) BayesCFMP, (B) CFMP, (C)BayesCFME.}
    \label{fig:experiment2}
\end{figure}

The results of the simulations conducted on two different settings, namely A and B, are exhibited in Figure \ref{fig:experiment} and Figure \ref{fig:experiment2} respectively. We refer the reader to Appendix \ref{app} for the details of the experimental settings considered, as well as additional experimental results for the two settings with different parameters. The mean and a 95\% CI (obtained from the variance estimate determined by the algorithms) are provided for every $\alpha$ for BayesCFMP, BayesRCFME and CFMP. The true expected ultimate reward is also displayed, which is unknown in real life applications. We have chosen two representative examples to illustrate how BayesCFME rectifies the deficiencies shown by CFMP and BayesRCFME by quantifying uncertainty corresponding to both data sets. All kernels considered were taken as RBF kernels.

In setting A  (Figure \ref{fig:experiment}), the true $\eta_\alpha$ lies outside the 95\% confident region for BayesRCFME for values of $\alpha$ close to 0. CFMP estimates suffer as well for low values of $\alpha$. Nonetheless, BayesCFMP uncertainty estimates seem to be best calibrated. In setting B (Figure \ref{fig:experiment2}), the true $\eta_\alpha$ lies outside 95\% CI provided by BayesCFME and CFMP in most of the $\Lambda = [-8, 8]$ considered. The uncertainty quantification seems to be well calibrated when considering the two sources of uncertainty through BayesCFMP.

\section{Conclusion and future work}

We have presented a Bayesian approach for the counterfactual distribution estimation problem based on Bayesian conditional mean embeddings. Such approach may be combined with the estimation of an unknown function in order to yield uncertainty estimates for the expectation of such function with respect to the counterfactual distribution. Building on the ideas presented in \cite{bayesimp}, we have proposed three methods for such task which differ in the sources of uncertainty considered. Furthermore, these techniques may also be applied to the off-policy evaluation framework, which can be seen as a generalization of the aforementioned problem. We have empirically compared the three proposed methods in two experimental settings, showing the value of considering the uncertainty stemming from the two different sources of data.

A line of research that could naturally follow this work is the exploration of a Bayesian kernel approach for distributional treatment effects, where the distribution of the treated and untreated has to be accounted for. This would open the door for two sample Bayesian tests for distributional treatment effects. A `functional' treatment effect, where the expectation would be taken over a (potentially estimated) function of the treatment effect, could also be addressed using such methodology. Moreover, studying closed-form solutions or approximations for nuclear dominant kernels other than Gaussian would open the door for doing model selection in the framework proposed, with the consequent computational and theoretical advantages.

\section*{Acknowledgements}

Diego Martinez-Taboada gratefully acknowledges the support provided by the Barrie Foundation.

\bibliographystyle{unsrtnat}
\bibliography{references}

\clearpage

\appendix
\section{Experimenal settings} \label{app}

In order to explore the performance of the proposed algorithms, two experimental settings are considered. For each of the examples, we work with two generated data sets $D_1 = \{ (u_i, a_i, r_i) \}_{i = 1}^N$ and $D_2 = \{ (\tilde{r}_j, y_j) \}_{j = 1}^M$. We then propose a set of new policies $\mathbb{P}_{A | U}^{\alpha}(a|u)$, where $\alpha \in \Lambda \subset \mathbb{R}$. We assume that $\mathbb{P}_U^\alpha(u)$ is invariant with respect to the historic data, thus we keep the given data $\bf{u}$ for the new policies (usual setting). The goal is to estimate $\eta_\alpha$, and to provide with uncertainty estimates accounting for the reliability of the measurement. All kernels considered were taken as RBF kernels. The two settings are defined such that:
\begin{enumerate}

\item [i.] Setting A:
\begin{itemize}
    \item $u_i \sim \mathcal{N}(5, 2)$.
    \item $\tilde{r_j} \sim Uniform(-2, 2)$.
    \item $a_i | u_i \sim 4 \mathcal{N}(0, 1)$.
    \item $r_i | u_i, a_i \sim (u_i + a_i) + 0.05 \mathcal{N}(0, 1)$
    \item $y_j | \tilde{r}_j \sim 2\sin{(\tilde{r}_j / 2)} + 0.05 \mathcal{N}(0, 1)$.
    \item $\mathbb{P}_{A | U}^\alpha(a | u) \sim  \alpha u + 0.05 \mathcal{N}(0, 1)$.
\end{itemize}

\item [ii.] Setting B:
\begin{itemize}
    \item $u_i \sim \mathcal{N}(5, 2)$.
    \item $\tilde{r_j} \sim Uniform(-2, 2)$.
    \item $a_i | u_i \sim  3 \mathcal{N}(0, 1)$.
    \item $r_i | u_i, a_i \sim \cos{u_i} + \sin{a_i} + 0.5 \mathcal{N}(0, 1)$
    \item $y_j | \tilde{r}_j \sim 1.5\sin{\tilde{r}_j} + 0.2 \mathcal{N}(0, 1)$.
    \item $\mathbb{P}_{A | U}^\alpha(a | u) \sim  2 \sin{\alpha} + 0.5 \mathcal{N}(0, 1)$.
\end{itemize}
\end{enumerate}

For both settings, we consider the parameters:
\begin{itemize}
    \item $N \in \{ 100, 200 \}$.
    \item $M \in \{ 100, 200 \}$.
\end{itemize}

\begin{figure}[t]
    \centering 
    \includegraphics[width=\linewidth]{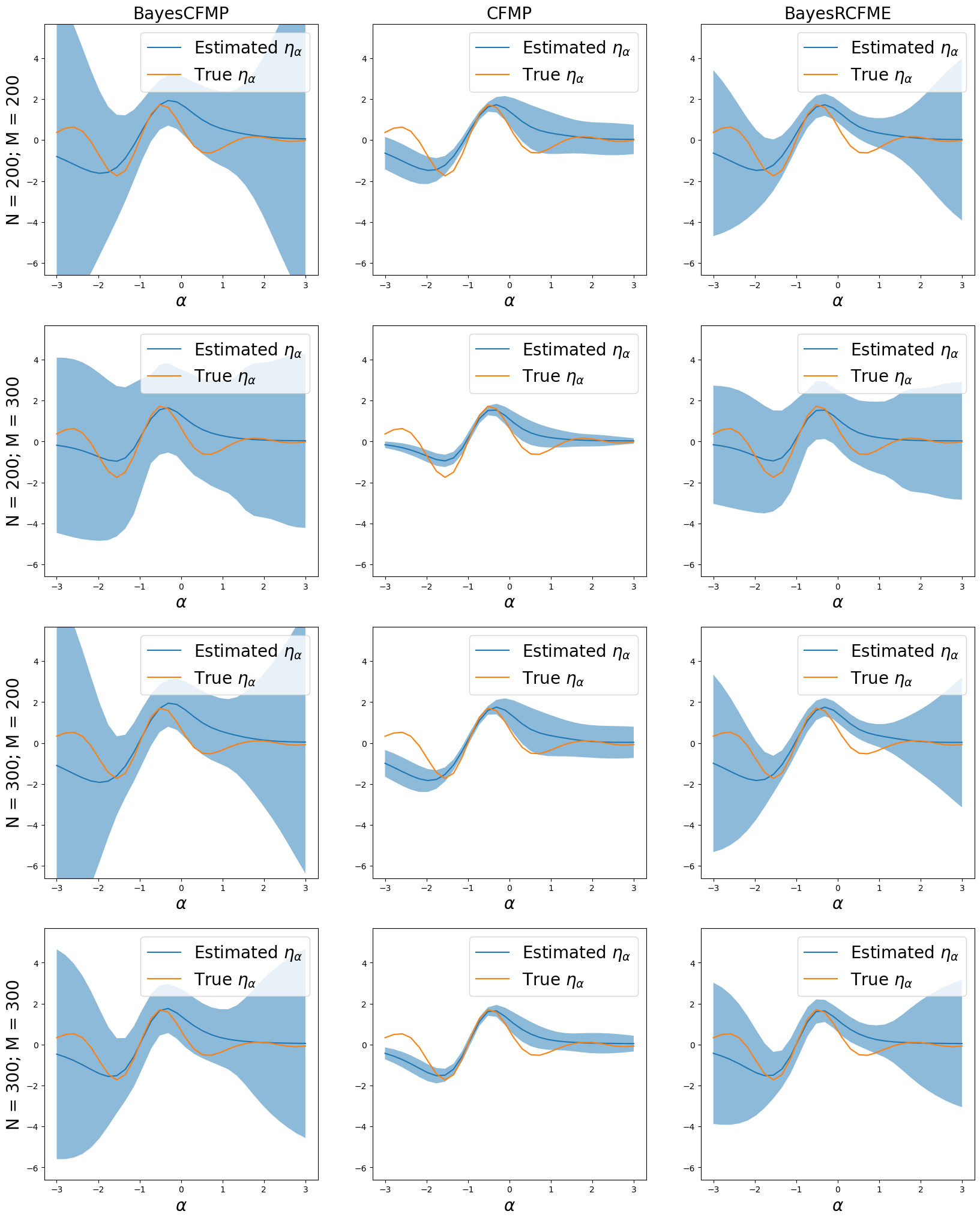} 
    \caption{Illustration of the experiment conducted on setting A. The mean and uncertainty estimates (95\% CI) of $\eta_\alpha$ are displayed for BayesCFMP, CFMP, and BayesCFME.}
    \label{fig:settingA}
\end{figure}

\begin{figure}[t]
    \centering 
    \includegraphics[width=\linewidth]{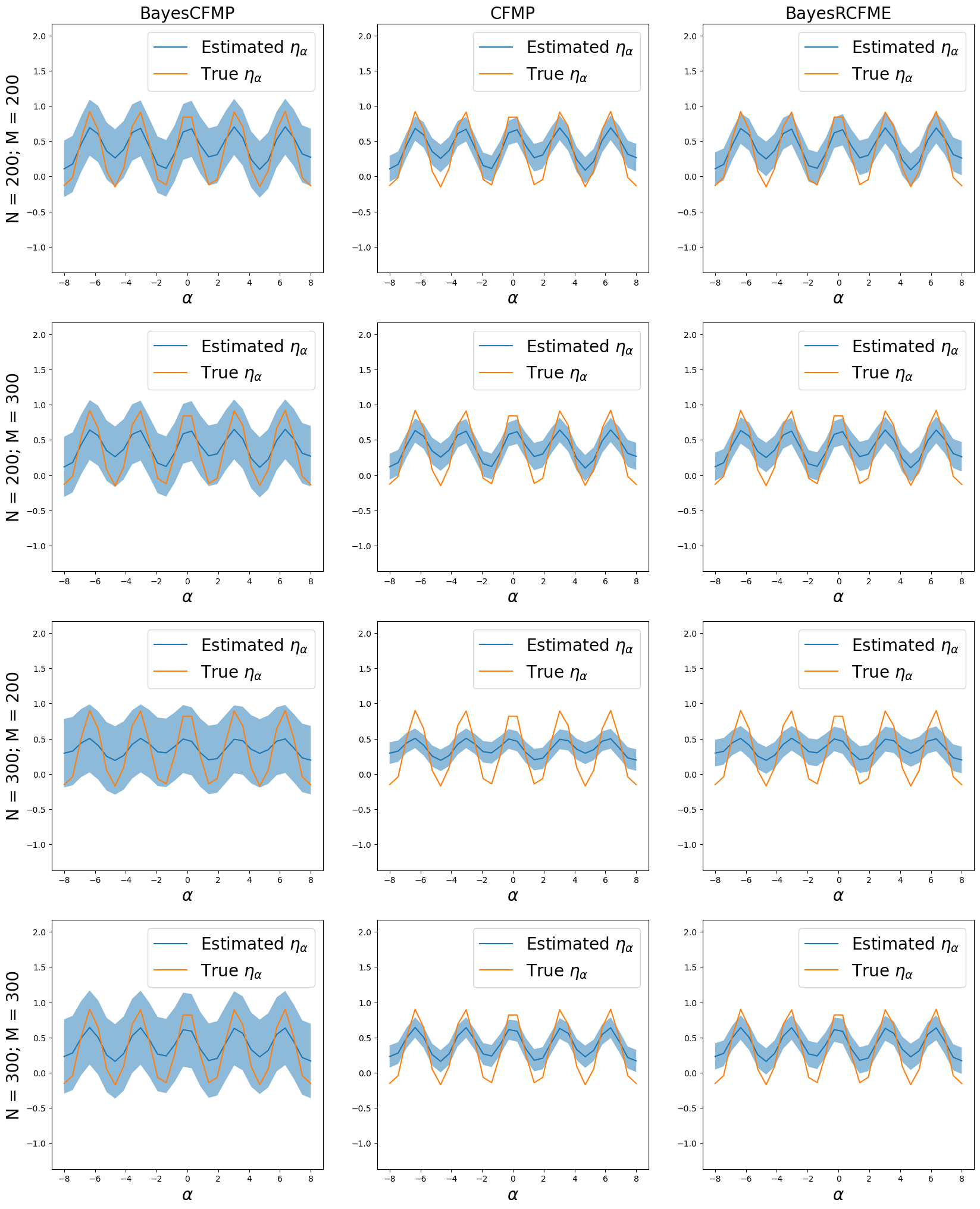} 
    \caption{Illustration of the experiment conducted on setting A. The mean and uncertainty estimates (95\% CI) of $\eta_\alpha$ are displayed for BayesCFMP, CFMP, and BayesCFME.}
    \label{fig:settingB}
\end{figure}

We exhibit the results of the experiments corresponding to Setting A and Setting B in Figure \ref{fig:settingA} and Figure \ref{fig:settingB} respectively.

\end{document}